\documentclass[sigconf]{acmart}
\usepackage{multirow}
\usepackage{pifont, paralist}
\usepackage{subcaption,graphicx}

\AtBeginDocument{%
  \providecommand\BibTeX{{%
    \normalfont B\kern-0.5em{\scshape i\kern-0.25em b}\kern-0.8em\TeX}}}

\setcopyright{acmlicensed}
\copyrightyear{2024}
\acmYear{2024}
\setcopyright{acmlicensed}
\acmConference[CIKM '24]{Proceedings of the 33rd ACM International Conference on Information and Knowledge Management}{October 21--25, 2024}{Boise, ID, USA}
\acmBooktitle{Proceedings of the 33rd ACM International Conference on Information and Knowledge Management (CIKM '24), October 21--25, 2024, Boise, ID, USA}
\acmDOI{10.1145/3627673.3680083}
\acmISBN{979-8-4007-0436-9/24/10}
\begin{document}

\title[COKE: Causal Discovery with Chronological Order and Expert Knowledge in High Proportion of Missing Manufacturing Data]{COKE: Causal Discovery with Chronological Order and Expert Knowledge in High Proportion of Missing Manufacturing Data}

% \affiliation{%

\author{Ting-Yun Ou}
\affiliation{%
  \institution{Department of Computer Science, National Yang Ming Chiao Tung University}
  \city{Hsinchu}
  \country{Taiwan}}
\email{outingyun.cs11@nycu.edu.tw}

\author{Ching Chang}
\affiliation{%
  \institution{Department of Computer Science, National Yang Ming Chiao Tung University}
  \city{Hsinchu}
  \country{Taiwan}}
\email{blacksnail789521.cs10@nycu.edu.tw}

\author{Wen-Chih Peng}
\affiliation{%
  \institution{Department of Computer Science, National Yang Ming Chiao Tung University}
  \city{Hsinchu}
  \country{Taiwan}}
\email{wcpeng@cs.nycu.edu.tw}

\renewcommand{\shortauthors}{Ting-Yun Ou, Ching Chang, and Wen-Chih Peng}

\begin{abstract}
% ###
Understanding causal relationships between machines is crucial for fault diagnosis and optimization in manufacturing processes. 
Real-world datasets frequently exhibit up to 90\% missing data and high dimensionality from hundreds of sensors. 
These datasets also include domain-specific expert knowledge and chronological order information, reflecting the recording order across different machines, which is pivotal for discerning causal relationships within the manufacturing data.
% Understanding causal relationships between machines in contemporary manufacturing processes is critical for efficient fault diagnosis and process optimization.
% In real-world scenarios, datasets often contain large amounts of missing data, sometimes as much as 90\%, as well as data from hundreds of sensors, resulting in high dimensionality. 
% Additionally, these datasets include domain-specific knowledge and time-sequence that reflect the order in which data was recorded across different machines.
% Because the knowledge within the manufacturing process influences the internal causal relationships, it can assist in establishing a causal relationship from the manufacturing dataset.
However, previous methods for handling missing data in scenarios akin to real-world conditions have not been able to effectively utilize expert knowledge.
Conversely, prior methods that can incorporate expert knowledge struggle with datasets that exhibit missing values. 
% Therefore, we propose \textcolor{red}{model name} to construct the causal graphs by leveraging expert knowledge and chronological order information among sensors in the manufacturing dataset. 
Therefore, we propose COKE to construct causal graphs in manufacturing datasets by leveraging expert knowledge and chronological order among sensors without imputing missing data.
% Without imputing missing values, we capitalize on the attributes of recipes within the manufacturing data to maximize the utilization of samples with missing values. 
Utilizing the characteristics of the recipe, we maximize the use of samples with missing values, derive embeddings from intersections with an initial graph that incorporates expert knowledge and chronological order, and create a sensor ordering graph.
The graph-generating process has been optimized by an actor-critic architecture to obtain a final graph that has a maximum reward.
% We propose MissGARL, a novel method for constructing causal graphs in manufacturing datasets by leveraging expert knowledge and chronological order among sensors, without imputing missing data. Utilizing recipe attributes, we maximize the use of samples with missing values, derive embeddings from intersections with an initial graph that incorporates expert knowledge, and create a sensor ordering graph. The graph-generating process, optimized via an actor-critic architecture, aims to produce a final graph with maximum reward.
Experimental evaluations in diverse settings of sensor quantities and missing proportions demonstrate that our approach compared with the benchmark methods shows an average improvement of 39.9\% in the F1-score.
Moreover, the F1-score improvement can reach 62.6\% when considering the configuration similar to real-world datasets, and 85.0\% in real-world semiconductor datasets.
The source code is available at \url{https://github.com/OuTingYun/COKE}.
\end{abstract}

\begin{CCSXML}
<ccs2012>
   <concept>
       <concept_id>10010147.10010178</concept_id>
       <concept_desc>Computing methodologies~Artificial intelligence</concept_desc>
       <concept_significance>500</concept_significance>
       </concept>
   <concept>
       <concept_id>10010147.10010178.10010187.10010192</concept_id>
       <concept_desc>Computing methodologies~Causal reasoning and diagnostics</concept_desc>
       <concept_significance>500</concept_significance>
       </concept>
   <concept>
       <concept_id>10010405.10010481.10010482</concept_id>
       <concept_desc>Applied computing~Industry and manufacturing</concept_desc>
       <concept_significance>500</concept_significance>
       </concept>
 </ccs2012>
\end{CCSXML}

\ccsdesc[500]{Computing methodologies~Artificial intelligence}
\ccsdesc[500]{Applied computing~Industry and manufacturing}

\keywords{Causal Discovery, Missing Data, Expert Knowledge, Chronological Order, Manufacturing Characteristics}



\maketitle
\section{Introduction}
\begin{figure}[h]

  \centering
  \includegraphics[width=0.7\linewidth]{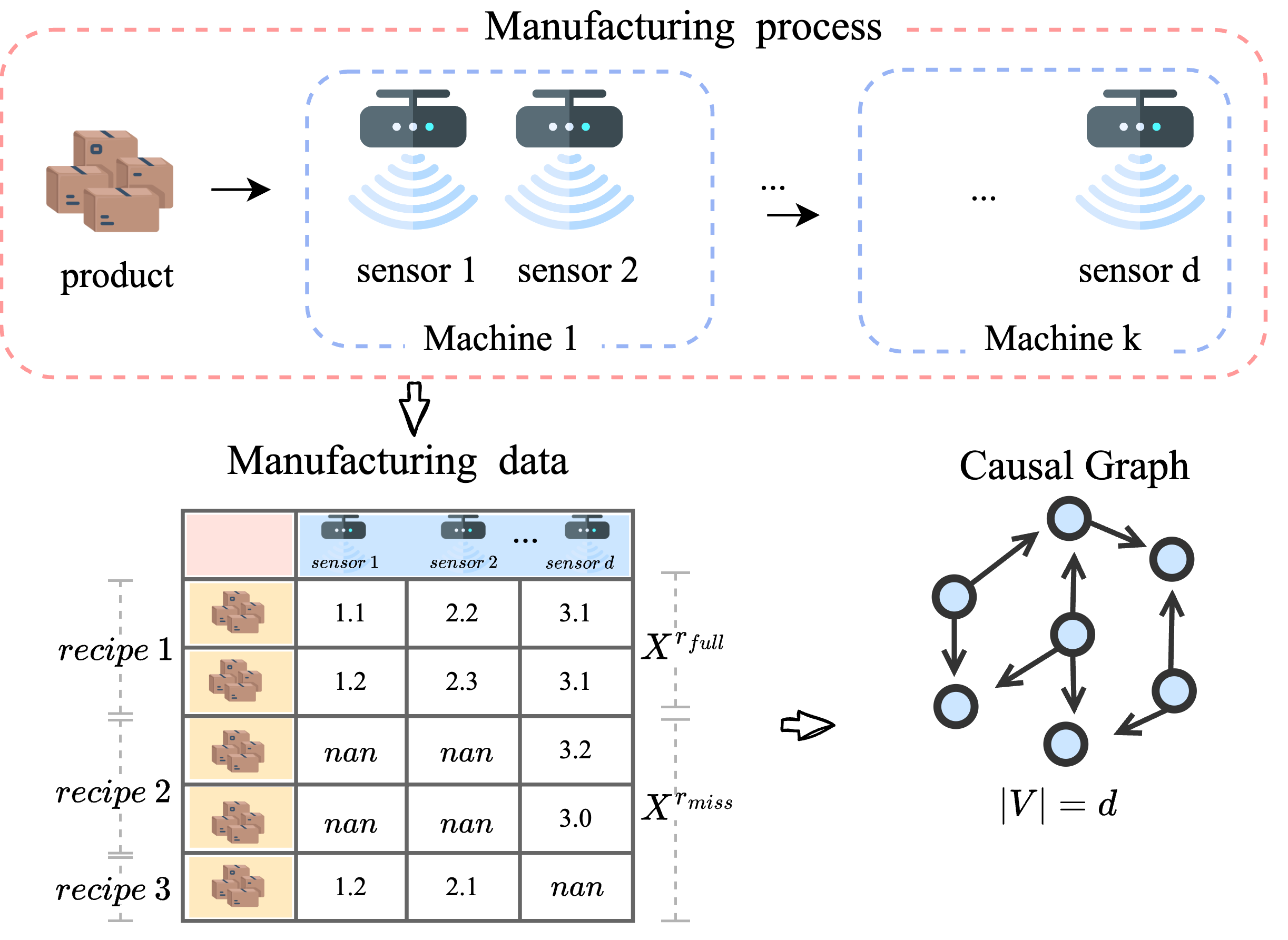}
  \caption{
  Manufacturing process workflow.
  Products sequentially interact with sensors across various machines, with an established order that ensures each sensor influences only subsequent sensors on the same or next machines. 
  % However, the presence of multiple recipes leads to missing sensor data and high rates of missing values within the dataset.
  }
  \vspace{-10pt}
  \label{fig:Motivation}
\end{figure}
% Paragraph1: Motivation
The manufacturing industry is increasingly transitioning towards automation and complexity to enhance product quality and variety \cite{mf-complexity,mf-auto}. 
For instance, semiconductor \cite{HECSL}  and fluid catalytic cracking processes \cite{mf-fluid} involve hundreds of machines. 
This complexity poses challenges in understanding the underlying causal mechanisms of production lines and identifying the root causes of system failures and product defects.
In this scenario, understanding the causal relationships between machines is crucial.
It enhances engineers' understanding of the system and enables the tracking of the root causes of failure events, thereby facilitating real-time error correction even in the absence of on-site engineers \cite{mf-imp1,mf-rca1,mf-rca2}.
Furthermore, understanding the causal relationships between components can offer preemptive alerts for prospective errors to reduce assembly line downtime \cite{cd-app-car,cd-mfp}. 
Therefore, employing causal discovery methods in manufacturing data is important in the recent evolution of manufacturing \cite{cdmf-md}.

%#### Paragraph2: Challenge1 - missing values problems
%#### Dataset including expert knowledge and chronology order which influence the causal relationship in the causal graph. It is important to consider information when constructing a graph. However, there are missing values in the dataset, which makes to unsuitable for applying exist method.

% Given the manufacturing dataset, these datasets incorporate domain-specific expert knowledge such as some edge must exist or be removed, and chronological order information which reflects the sequence of recordings across different machines. It is crucial for identifying causal relationships within the manufacturing data considering expert knowledge and chronological order information.

In the manufacturing process, the datasets include domain-specific expert knowledge to modify or retain certain edges and chronological order capturing the sequence of data recordings across various machines.
The expert knowledge and chronological order reflect true interactions and dependencies influenced by the manufacturing process  \cite{cdmf-md}. 
Expert knowledge distinguishes between causal relationships from mere correlations, while chronological data captures the sequence of events, aiding in identifying how changes or errors propagate through the system.
% Expert knowledge can distinguish between mere correlations and genuine causal relationships that are operationally significant.
% Manufacturing processes are sequential, with outputs from one stage often as inputs to the next. 
% Chronological data captures this sequence, enabling the identification of how changes or errors propagate through the system.
Therefore, it is essential to leverage information when accurately identifying causal relationships within the manufacturing data.
% we can establish a causal graph through causal discovery methods. % Traditional methods include constraint-based \cite{PC,RFCI,PC-IR}, score-based \cite {GreedySearch,GES,DAGSP}, and functional causal model approaches \cite{CAM,NonCD,ICA-LiNGAM,CDCANM,IPNCD}. 
Recent research \cite{GARL} inspired by \citet{CAM}, divided causal discovery into two stages: ordering graph generation and neighbor variables selection.
This approach allows the design of a model that can utilize a prior knowledge graph for constructing the causal graph.
% However, in manufacturing data, the characteristic of recipes results in high-missing proportions values in the dataset, as shown in Figure \ref{fig:Motivation}.
However, in manufacturing data, the recipes result in datasets with a high proportion of missing values, as shown in Figure \ref{fig:Motivation}.
The recipes determine which machines and sensors are involved in the production process, and the sequence of machines used in it. 
The variability in recipes across different product configurations can lead to the occurrence of missing values in the dataset \cite{recipe2miss_1}. 
For example, consider a manufacturing facility that produces semiconductors with varying specifications. 
Certain product configurations may bypass certain machines or sensors, resulting in missing data for those particular sensors \cite{HECSL}. 
As a result, the dataset may exhibit a high missing rate, particularly for sensors associated with less frequently used product configurations. 
Consequently, before applying a method \cite{GARL} that could effectively leverage expert knowledge and chronological order in manufacturing data, it is essential to impute missing data, potentially compromising the optimal efficacy of the solution.
%#### Paragraph3: Challenge 2 - cooperate expert knowledge
%#### The characteristic of recipes leads to a large proportion of missing values in manufacturing data, resulting in unsuitable existing causal discoveries for missing data methods. Also, these methods can not consider expert knowledge and chronological information when graphing construction. 
% Therefore, to utilize graph construction methods, it becomes necessary to impute these missing values to achieve a complete dataset.
% The characteristic of recipes in manufacturing data often results in a large proportion of missing values (up to 90\%), necessitating the imputation of these values to achieve a complete dataset before employing graph construction methods.

With missing rates in manufacturing data reaching up to 90\%, imputed data significantly impact the dataset's distribution and affect the performance of results.
To avoid bias from imputed data, some previous approaches will pre-define the types of missing data. 
According to \cite{IMS}, based on the different types of missingness, the underlying missing mechanism can be divided into three basic types: missing at random (MAR), missing completely at random (MCAR), and missing not at random (MNAR). 
% MAR occurs when missingness depends solely on observed variables; MCAR arises when missingness is uniform across all data; MNAR indicates varying missingness probabilities due to unknown factors.
However, the missingness in manufacturing data stems from variations in recipes, leading to unsuitable accurate definitions of the types of missing data.
This obstacle complicates the selection of appropriate methods for causal discovery and the optimization of their utilization in data processing.
% Furthermore, existing methods addressing missing data struggle to leverage expert knowledge and chronological order information among sensors in manufacturing datasets for effective assistance.
Other methods that do not specify missing data types struggle with incorporating expert knowledge and chronological order information among sensors \cite{VISL,mvpc}, and are inefficient in processing datasets with hundreds of variables \cite{missdag}. 
% Although there are some existing methods\cite{missdag, VISL,mvpc} addressing missing data, they struggle to leverage expert knowledge and chronological order information among sensors in manufacturing datasets for effective assistance.

% ###
%### Briefly talk about the challenge of paragraph 2 and paragraph 3
When constructing a causal graph from manufacturing data, expert knowledge and chronological order information are crucial as they impact the causal relationships between sensors. 
However, existing methods struggle to address the challenge posed by high proportions of missing values in data with varied recipes. 
Additionally, methods capable of handling missing data fail to effectively leverage expert knowledge and chronological order information.
%### Our framework
To address these challenges, we propose COKE, which leverages \textbf{C}hronological \textbf{O}rder and expert \textbf{K}nowledg\textbf{E} using a graph attention network. This approach optimizes the utilization of samples with missing values by recipes in manufacturing data, thereby avoiding the need for data imputation and achieving accurate variable embeddings.
Upon obtaining embeddings for each variable (sensor), we use a decoder to sequence the variables by causal order. 
We then remove edges from the ordering graph based on these sequenced variables using the initial graph.
% Our model adopts an actor-critic architecture from embedding acquisition through variable ordering. 
% We employ the Bayesian Information Criterion (BIC) score as a reward function to optimize our graph generation model, aiming to maximize this reward. 
We employ an actor-critic architecture to optimize our graph generation model, aiming to maximize the reward by minimizing the Bayesian Information Criterion (BIC) score.
Throughout the training, we record the graph in each iteration, ultimately selecting the graph with the highest reward as the final causal graph for the manufacturing dataset.

Our contributions can be summarized as follows: 
1) We introduce COKE, a novel deep learning approach designed to address the challenge of effectively utilizing expert knowledge and chronological order in datasets with up to 90\% of missing values. 
2) We utilize recipes in the manufacturing process to generate embeddings that comprehensively represent variables without imputing missing values. 
3) In the scenarios involving diverse quantities and missing rates on synthetic data simulating the real-world dataset distributions, our approach demonstrates enhancing the F1-score by 39.9\% compared to the benchmark. Additionally, on real-world data, it achieves an improvement above 85.0\%. Furthermore, through ablation studies, we demonstrate the impact of expert knowledge and chronological order in the graph-constructing process from manufacturing data.
%    Additionally, we use a graph attention network to leverage these embeddings and initial graphs with expert knowledge and chronological order from the manufacturing data.  
% \begin{compactitem}
%    \item We introduce COKE, a novel deep learning approach designed to address the challenge of effectively utilizing expert knowledge and chronological order in datasets with up to 90\% of missing values. 
%    \item We utilize recipes in the manufacturing process to generate embeddings that comprehensively represent variables without imputing missing values. 
%    Additionally, we use a graph attention network to leverage these embeddings and initial graphs with expert knowledge and chronological order from the manufacturing data.   
%    \item In the scenarios involving diverse quantities and missing rates on synthetic data simulating the real-world dataset distributions, our approach demonstrates enhancing the F1-score by 39.9\% compared to the benchmark. 
%    Additionally, on real-world data, it achieves an improvement above 85.0\%. 
%    Furthermore, through ablation studies, we demonstrate the impact of expert knowledge and chronological order in the graph-constructing process from manufacturing data.
% \end{compactitem}

\section{Related Work}
\subsection{Causal Discovery with Incomplete Data} 
Causal graph construction in the presence of missing data requires considering the impact of missing values on causal discovery. 
\citet{mvpc} extended the constraint-based PC algorithm by analyzing observed data to identify potential errors or inconsistencies in causal relationships between variables. 
They also proposed corrective measures to achieve a more accurate representation of the true causal graph. 
\citet{cd-miss2} utilized additive noise models, expanding the identifiability results of causal discovery methods, including the recognition of causal skeletons and weakly self-concealing missingness. 
However, these constraint-based methods based on the PC algorithm result in ineffectively constructing graphs for data with hundreds of variables because of the exponential growth of conditional sets \cite{fpc}.
MissDAG \cite{missdag} employs the Monte Carlo EM framework. 
In the E-step, the model estimates the data distribution from the observed available data, taking into account the missing values. 
The M-step then utilizes the causal discovery model to estimate the data and identify the causal graph among variables. 
% , which is represented as a directed acyclic graph.
However, this method is time-consuming for hundreds of variables and cannot utilize expert knowledge and chronological order in the graph-constructing process. 
Several methods use machine learning techniques to address the missing values in the dataset.
\cite{VISL, ICL, MICwgan,cd-rl-miss} obtained the precise characteristics of data distributions, enabling the identification of causal relationships by uncovering the correct distribution from the data.
These methods perform end-to-end training on generative adversarial network-based imputation models and causal structure learning models. 
% However, these methods are also incompatible considering other information from manufacturing datasets such as chronological order information.
However, they are unsuitable for manufacturing datasets since they do not account for expert knowledge and chronological order.

\subsection{Causal Discovery in Manufacturing Data}  
Advancements in manufacturing processes, automated measurement tools, and real-time data collection have improved data quality, facilitating accurate representation of product conditions \cite{cdmf-faultdetect}. 
As a result, there has been significant progress in applying causal discovery techniques within the manufacturing data. 
\citet{HECSL} proposed a causal structure learning method for wafer manufacturing data, which utilizes a hierarchical structure to decompose complex causal discovery into smaller substructures.
It integrates learning multiple causal discovery algorithms to improve the robustness and accuracy of the learned structure. 
% However, this work assumes that sensors at the same machines do not have causal relationships, so it cannot find accurate causal relationships between sensors in the same machine. 
However, this work assumes that sensors on the same machine lack causal relationships, thereby limiting its ability to accurately identify the causal-effect relationships between sensors within the same machine.
\citet{cdmf-md} applied data mining and knowledge discovery to root cause analysis in the manufacturing and interconnected industries, while \cite{cdmf-failure} constructs a causal graph taking the minimum description length (MDL) constructed the scoring function, followed by subsequent root cause analysis and application to process reactors. 
However, these methods only consider a few machines in the dataset, as does the work in multi-stage PCB manufacturing \cite{cdmf-dm}. 
% Methods proposed by \cite{cdmf-mdl} can be applied to a high proportion of missing values datasets, but they do not account for missing data. 
% Understanding hidden causal relationships in manufacturing processes can provide significant assistance. 
% For example,  \cite{cdmf-industrial} proposes a new hybrid method based on Transfer Entropy (TE), combining process connectivity information using an explicit search algorithm to detect causal relationships in an industrial cardboard machine.
% \citet{cd-app-car} demonstrates how causal relationship models can be incorporated into automotive production monitoring tools to aid in fault prevention. 
% The method proposed by \cite{mf-fluid} is applied to fluid catalytic cracking devices.
% ###
% Although numerous methods are available for constructing causal graphs from manufacturing data, they struggle with datasets characterized by high proportions of missing values or dimensions. 
% These problems lead to improper application of our problem.
While there are many methods for constructing causal graphs from manufacturing data, they face challenges with datasets in high proportions of missing values or limited dimensions, affecting improper application for our problem.
% The previous benchmark method, MissDAG \cite{missdag}, constructs datasets with missing values but lacks consideration of manufacturing data knowledge. 
% It also requires significant processing time due to hundreds of sensors. 
% The recent method for semiconductor dataset \cite{HECSL}, can handle a high proportion of missing data and numerous sensors. 
% However, its variable definition combines multiple sensors, preventing accurate detection of causal relationships between individual sensors.
% GARL \cite{GARL} generates a causal graph by considering the prior knowledge, but it cannot handle the missing values in the dataset. 

Despite the high proportion of missing values in the dataset, we can obtain complete datasets through imputation for causal discovery.
Notears \cite{notears}, a widely-used method, reformulates causal discovery as a continuous function optimization problem from a combination problem. 
This method inspired \cite{CAM,CORL,GARL,VISL} to integrate machine learning techniques for causal discovery problems.
ICA-LiNGAM \cite{ICA-LiNGAM} addresses linear non-Gaussian models with continuous-valued data, while GARL\cite{GARL}  trains the graph-generating process by optimizing the score function using an actor-critic architecture.
However, imputation bias with up to 90\% missing values leads to inaccurate results of causal discovery.
\section{Preliminary}
\subsection{Problem Formulation}
In the manufacturing dataset, the recorded process follows the description in Figure \ref{fig:Motivation}. 
The dataset is represented as $D = (X, d, k,  \Gamma)$, where $X=[X_{1},\dots, X_{d}]^T\in \mathbb{R}^{N \times d}$, with $N$ samples and $d$ sensors. 
$k$ denotes the number of machines and $k_{d}$ means the corresponding machine for sensor $d$.
In the graph generated from observational data, each variable $x_i$ corresponds to sensor $i$ in the manufacturing process. 
$\Gamma=\{r_{1}, \ldots, r_{s}\}$ represents a set of $s$ recipes within $X$, where each $r_{i}$ comprises a different number of samples denoted as $n_i$, resulting in a total of $N = \sum_{i=1}^{s} n_{i}$.
Then we can partition the dataset into $X=[X^{r_{1}}, \ldots, X^{r_{s}}]$ and each $X^{r_i} \in \mathbb{R}^{n_{i}\times d}$ contains $d$ variables. 
The dataset $X$ is divided by recipes into subsets $X^{r_{i}}$, each containing observed data $X^{r_{i}}_{o(r_{i})}$ and missing data $X^{r_{i}}_{m(r_{i})}$.
Here, $m(r_{i})$ denotes the set of variables (sensors) that are missing due to bypass in the recipe $r_i$, while $o(r_{i})$ represents the set of variables that are observed.
As shown in Figure \ref{fig:Motivation}, $o(r_{3})$ is $\{x_{1}, x_{2}, \dots\}$, and $X^{r_{3}}_{o(r_3)}$ denotes a complete dataset in $r_3$. 
We also define the specific recipe as $r_{full}$ where $|o(r_{full})| = d$ and the corresponding subset $X^{r_{full}}$ is completed.
Hence, there are $(s-1)$ subsets that are incomplete, denoted as $X^{r_{miss}} = \{ X^{r_{i}} \: | \: r_{i} \in \Gamma \:\text{and}\: |o(r_{i})| < d \}$. Example of $X^{r_{full}}$ and $X^{r_{miss}}$ are shown in Figure \ref{fig:Motivation}.

\subsection{Model Definition}
In this work, we represent the causal graph as ${G}=({V,E}) $ , where $V=\{1,...,d\}$ is a set containing $d$ variables, and $E=\{ (i,j)\:|\:i,j=1,\ldots d\}$ is a set of directed edges from variable $i$ to variable $j$.
Each variable in the causal graph is denoted as $x_{i}$, where $i \in V$.
The assumption of the data generation process is followed by 
\begin{equation*}
\small
\label{eq:data_gen}
    x_i = f_{i}(x_{Pa(i)})+ \eta_{i}, \:i=1,\ldots ,d,
\end{equation*}
where $x_{Pa(i)} = \{k\:|\:(k,i) \in E\}$ is the set of parent variables for $x_{i}$, and $f_{i}(\cdot)$ represents the causal relationship between $x_{i}$ and $x_{Pa(i)}$. 
We follow the assumption of \cite{CORL} that assumes the noise $\eta_{i} \sim \mathcal{N}(0,\sigma^2)$ is jointly independent with variables in ${x_{Pa(i)}}$ and the causal minimality for each function $f_{i}(\cdot)$ is not constant. 
Additionally, we assume that all variables are measurable (causally sufficient) and there are no confounders, which are latent causes of the observed variables in the dataset. 
Given the observational data $X={[X_{1},..X_{d}]}^T\in \mathbb{R}^{N \times d}$ the goal of causal discovery is to find hidden cause-effect relationships between variables from the observational data.

% The ANM can be identified within the Markov equivalence class, we focus on a specific class called linear Structural Equation Model (SEM):
% \begin{equation}
% \small
%     X_i = \sum_{k \in X_{Pa(i)}} f_{i}(X_k) + \eta_{i} \:\:\:\:(i=1,\ldots,d)
% \end{equation}
% the above model can be called Linear Non-Gaussian model with $f_{i,k}$ as linear and $n_{i}$ as a Gaussian model. This type of model can be identified by LiNGAM \cite{shimizu2014lingam}. Previous work also identifies linear SEM when facing the Linear Gaussian model and non-linear Gaussian model. 
% ###

\subsection{Ordering Variable Definition}
In previous work, the task of identifying a direct acyclic graph can be defined as determining a variable ordering \cite{CAM, CORL, Order-BS}. 
Let $\Pi$ be a set of ordering variable (sensors) for $V$ and $|\Pi|=d$, if the $q$-th variable in $\Pi$ is $x_{i}$ then  $\Pi(q) = x_{i}$. 
We denote a direct acyclic graph (DAG) as $\mathcal{G}$.
For any ordering variable set $\Pi$, we can construct a  DAG $\mathcal{G}^{\Pi}$ which has directed edges from $\Pi(i)$ to $\Pi(j)$ for any $i<j$, as shown in Figure \ref{fig:multiply}. 
A $\mathcal{G}$ can consist of more than one ordering variable set, denoted as:
\begin{equation*}
\small
\Phi(\mathcal{G}) = \{ \Pi : \mathcal{G}^{\Pi} \: \text{is a super-DAG of} \:\: \mathcal{G}\},
\end{equation*}
where a super-DAG of $ \mathcal{G}$ is a DAG whose edge set is a superset of $\mathcal{G}$. 
To find true DAG $\mathcal{G}^{*}$, \citet{CAM} separated the causal discovery problem to find the ordering variable sets of the true DAG $\Phi(\mathcal{G^{*}})$ and variables selection in each $\Pi \in \Phi(\mathcal{G^{*}})$.
We constructed an initial graph $ g_{init}$ with $d$ variables to assist the variables selection in each ordering variable set in $\Phi(\mathcal{G^{*}})$. 
To ensure comprehensive consideration of information between variables, the graph contains all potential causal relationships; thus, all variables are fully connected in the initial graph $g_{init}$.

\section{Method}
\begin{figure*}
  \centering
  \includegraphics[width=0.8\linewidth]{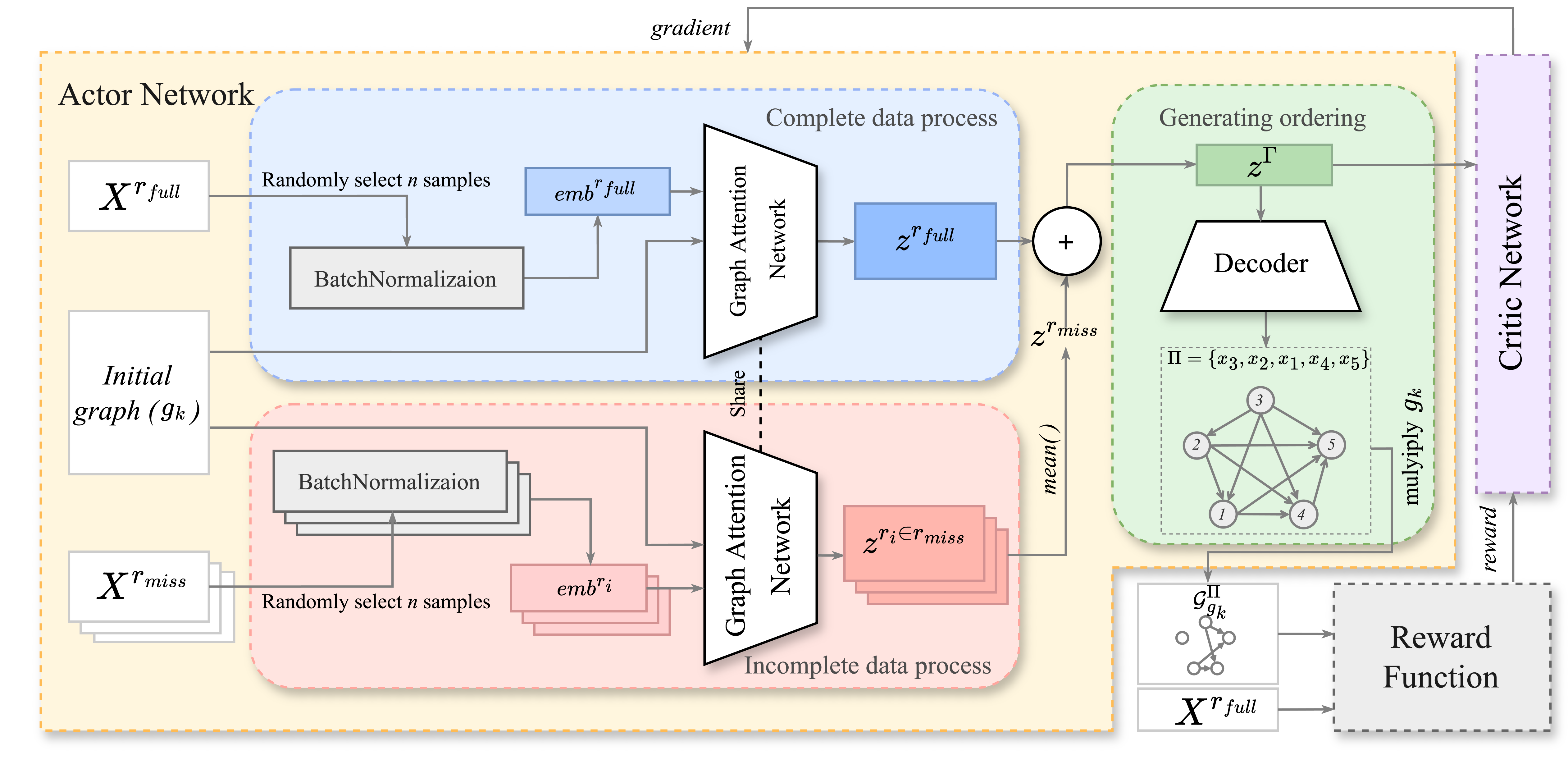}
  \caption{
  The overview framework of COKE.
  In this framework: 
  1) the actor obtains a causal graph from the ordering graph, which is generated by the dataset and the initial graph with expert knowledge and chronological order information; 
  2) the reward function evaluates the graph produced by the actor using the complete dataset; 
  3) the critic network assesses the current state's value, guiding the actor network's update through gradients derived from the reward and value.}
  % \vspace{-10pt}
  \vspace{-0.4cm}
  \label{fig:framework}
\end{figure*}
\label{sec:method}
In our task, we aim to derive the true DAG $\mathcal{G}^*$ from the observed dataset $X$, which contains a high proportion of missing values. 
To address this challenge, we break down the causal discovery task into two main components: ordering graph generation and variables selection. 
Firstly, in Section \ref{sec:initial-graph}, we modify an initial graph by incorporating expert knowledge and chronological ordering. 
In Section \ref{sec:ordering_graph_generating}, we generate the ordering without imputation by leveraging the characteristics of recipes in the manufacturing dataset and utilizing the initial graph. 
Then, in Section \ref{sec:optimizing_rl}, we determine the ordering variable $\Pi^*$ that maximizes the reward function through variables selection based on the initial graph. 
The framework of our proposed method is illustrated in Figure \ref{fig:framework}. 
\subsection{Initial Graph Construction with Expert Knowledge and Chronological Order}
\label{sec:initial-graph}
In manufacturing data, since products pass through machines sequentially,  chronological order occurs between the machines. 
Therefore, we reduce the space of $\Phi(\mathcal{G^*})$ by permuting the machines in the manufacturing data: 
\begin{equation}
\small
\bar{\Phi}(\mathcal{G^*}) = \{ \Pi\:|\: for \:i<j ,\:\: k_{\Pi(i)} \leq k_{\Pi(j)}\} \subset \Phi(\mathcal{G^*}),
\end{equation}
where $k_{\Pi(i)}$ represents the corresponding machine of sensor ${\Pi(i)}$. 
For true DAG, $\bar{\Phi}(\mathcal{G^*})$ contains ordering variable sets that conform to chronological order information. 
We refine the initial graph $g_{init}$ by removing the edges that are not in the union of all $\Pi \in \bar{\Phi}(\mathcal{G^*})$. 
Then, we define $g_k$ as the initial graph that accounts for the chronological order among machines.
We apply preliminary neighbor selection (PNS) to the initial graph $g_{k}$ by the assumption that the result of PNS on  $g_{k}$ contains all correct edges \cite{CAM}.
Subsequently, we modify the edges based on expert knowledge by removing or adding them accordingly.
% Therefore, obtaining the initial graph $g_k$ contains expert knowledge and chronological order information for ordering graph generating and variable selection in Section \ref{sec:ordering_graph_generating}.
Therefore, the initial graph \( g_k \) includes expert knowledge and chronological order information, making it ready for ordering graph generation and variables selection.

\subsection{Enhancing Ordering Graph Generation with Recipe Characteristics}
 %  and Recipes
\label{sec:ordering_graph_generating}
In each training iteration, $n$ samples are randomly selected for each $X^{r_i}$ to ensure uniform subset sizes. 
If the number of samples in $X^{r_i}$ is less than $n$, then $r_i$ is excluded from the training process. 
It is assumed in the subsequent section that all subsets contain more samples than $n$.

% ###
\subparagraph{\textbf{Complete Data Process.}}
% We start the process by using batch normalization on $X^{r_{full}}$ to get initial embeddings $emb^{r_{full}}$ for each variable. 
% That is, for each $i \in V$, $emb^{r_{full}}_i = BatchNorm(X^{r_{full}}_i)$.
% Then applying graph attention networks (GAT) to update the $emb^{r_{full}}$ by considering its parent variables in the $g_{k}$:
% \begin{equation}
% emb_{Pa(j)} = \{emb^{r_{full}}_i\:|\: i\in x_{Pa_{g_k}}(j)\},
% \end{equation}
% \begin{equation}
% z^{r_{full}}_{j}= GAT(emb_{Pa(j)}), 
% \end{equation}
% where $z^{r_{full}}$ represents an embedding that includes expert knowledge and chronological order information from complete data.

We start the process by using batch normalization on $X^{r_{full}}$ to get initial embeddings for each variable and then apply graph attention networks (GAT) to update the embeddings by considering its parent variables in the $g_{k}$:
\begin{equation}
z_{Pa(j)} = \{BatchNorm(X^{r_{full}}_i) \mid i \in x_{Pa_{g_k}}(j)\},
\end{equation}
\begin{equation}
z^{r_{full}}_j = GAT(z_{Pa(j)}),
\end{equation}
where $z^{r_{full}}$ represents an embedding that includes expert knowledge and chronological order information from the complete data.

\subparagraph{\textbf{Incomplete Data Process.}}
Since 90\% of samples with missing values are not utilized, we use the characteristics of recipes in $X^{r_{miss}}$ to obtain more accurate embeddings. 
For each $r_{i} \in r_{miss}$, the corresponding subset $X^{r_{i}}_{o(r_{i})}$ is individually processed by batch normalization, producing initial embedding for the observation variables.
% $z^{r_{i}}_{o(r_{i})}$ for observations variables: 
% \begin{equation}
% z^{r_{i}}_j = BatchNorm(X^{r_{i}}_j), for \: x_{j}\in o(r_{i}).
% \end{equation}
Subsequently, initial embedding is passed through the GAT used in the complete data process for taking into account the embedding of parent variables in $g_{k} $to update each embedding:
\begin{equation}
\label{eq:intersection}
z^{r_{i}}_{Pa(j)} = \{BatchNorm(X^{r_{i}}_l)|\: l\in \{x_{Pa_{g_k}}(j) \cap o(r_{i}) \}\},
\end{equation}
\begin{equation}
\label{eq:incomp}
z^{(r_i,\:g_k)}_{j}= GAT(z^{r_{i}}_{Pa(j)}), 
\end{equation}
where we derive embedding $z^{(r_i,\:g_k)}$ from the intersection of the $g_{k}$ and the observed variables in $o(r_{i})$. 
In Equation (\ref{eq:intersection}), for each variable $x_{j} \in o(r_{i})$ in $g_{k}$, if there exists a parent variable $x_{l}$ that is not observed in $r_{i}$, that is $x_{l} \in x_{Pa_{g_k}(j)}$ and $x_{l}\in m(r_{i})$, we exclude $x_{l}$ from consideration when updating the initial embedding of $x^{r_i}_j$ in $r_{i}$.
Each variable has a distinct number of embeddings as each recipe contains varying observed variables, leading to differing observation frequencies in different variables. 
Therefore, we average all embeddings that consider knowledge from the initial graph for each variable:
\begin{equation}
\small
    z^{r_{miss}}_j = Mean(\{z^{(r_i,\:g_k)}_{j}
    \: | \: r_{i} \in r_{miss} \:\text{and}\: j\in o(r_{i})\}), 
\end{equation}
where $z^{r_{miss}}$ is an embedding from the incomplete data process.

% After obtaining the embeddings from the complete data process and incomplete data process, we add these embeddings with trainable parameters:
After obtaining the embeddings from the complete and incomplete data processes, we use trainable parameters \(\theta_{full}\) and \(\theta_{miss}\) to control the influence of $z^{r_{full}}$ and $z^{r_{miss}}$ on the variables utilized for ordering generation:
\begin{equation}
    z^{\Gamma} = \theta_{full} \cdot z^{r_{full}} + \theta_{miss} \cdot z^{r_{miss}},
\end{equation}
where $z^{\Gamma}$ is the final embedding of the variable.
The model will autonomously adapt the proportion of $z^{r_{full}}$ and $z^{r_{miss}}$ based on the reward of the final graph. 

\begin{figure}[h]
  \centering 
  \setlength{\abovecaptionskip}{3pt}
  \includegraphics[width=0.6\linewidth]{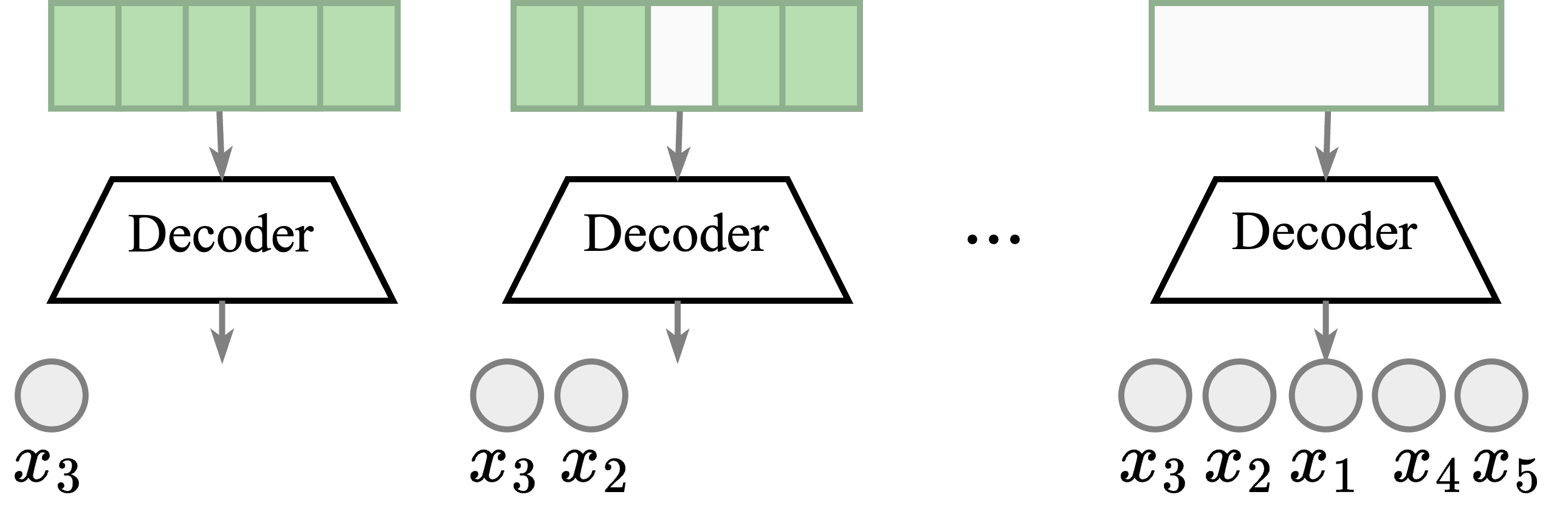}
  \caption{Generation of the ordering variable \(\Pi\).}
  \vspace{-0.3cm}
\label{fig:gen-order}
\end{figure}

\begin{figure}
\vspace{-0.3cm}
  \centering 
  \setlength{\abovecaptionskip}{4pt}
  \includegraphics[width=0.8\linewidth]{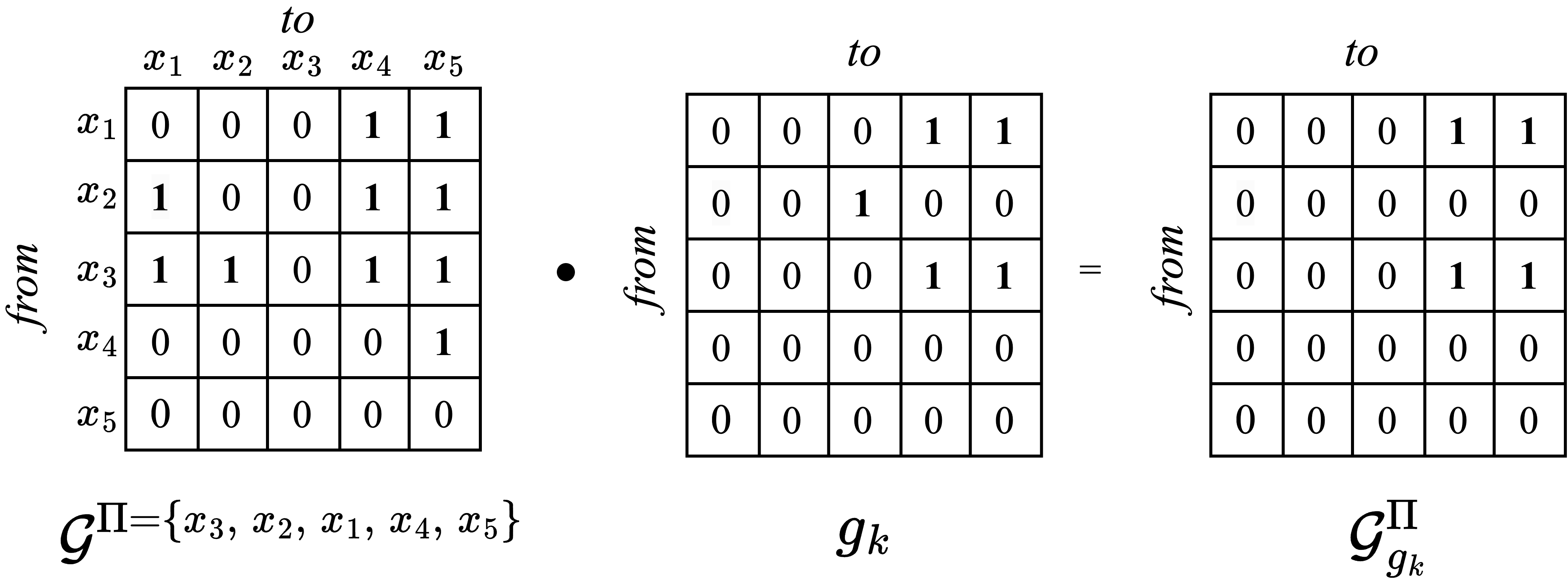}
  \caption{The transformation from $\mathcal{G}^{\Pi}$ to  $\mathcal{G}^{\Pi}_{g_{k}}$}
  \vspace{-0.6cm}
\label{fig:multiply}
\end{figure}

\subparagraph{\textbf{Generating Ordering.}} 
We derive the ordering variable $\Pi$ from a decoder. 
During each generative iteration as shown in Figure \ref{fig:gen-order}, we produce one variable until all variables have been generated, and the resulting sequence is defined as the ordering variable set. 
To prevent the embeddings of previously generated variables from affecting subsequent variable selection, we mask embeddings that have been generated. 
We get the embedding in the $t$-th generative iteration by masking previous variables in $\Pi$:
% \begin{equation}
% \label{sen:masking}
% \small
% Mask(emb^\Gamma,t) =\{ emb^\Gamma_i \:|\: emb^\Gamma_i = 0 \:\:\text{if}\:x_{i} \in \{\Pi(p)\ | 1< \:p \leq t\}\},
% \end{equation}
% \begin{equation}
% \label{eq:masking}
% \text{Mask}(z^\Gamma, t) = 
% \begin{cases} 
% 0 & \text{if } x_i \in \{\Pi(p) \mid 1 \leq p \leq t\} \\
% z^\Gamma_i & \text{otherwise}
% \end{cases},
% \end{equation}
\begin{equation}
\label{eq:masking}
\small
\text{Mask}(z^\Gamma, t) = \{ z^\Gamma_i \}, \quad \text{where} \quad z^\Gamma_i = \begin{cases} 
0 & \text{if } x_i \in \{\Pi(p) \mid 1 \leq p \leq t\} \\
z^\Gamma_i & \text{otherwise}
\end{cases},
\end{equation}
and then we generate the next variable by:
\begin{equation}
\small
\Pi(t+1) = Decoder(Mask(z^\Gamma,t)).
\end{equation}
Generative iteration is repeated $d$ times to obtain the ordering variable set. 
We denote $\Pi_{t}$ as the $t$-th ordering variable set generated by the training process. 
In the manufacturing data, the  $\mathcal{G}^\Pi_{g_k}$ obtained by multiplying $g_{k}$ and $\mathcal{G}^\Pi$ for each training iteration considering the knowledge in $g_k$ is shown in Figure \ref{fig:multiply}.
Upon completing the whole training process, we obtain the graph $\mathcal{G}^* \in \{\:\mathcal{G}^{\Pi_{i}}_{g_k} \:|\: i=0 \dots,t\:\}$ that maximizes the reward function as the causal graph from the observational dataset.

\subsection{Reinforcement Learning Architecture}
\label{sec:optimizing_rl}
We utilize an actor-critic framework to train the ordering graph generation process, conceptualizing it as a multi-step decision-making process \cite{CORL}. 
During each training iteration, embeddings are obtained by resampling both complete and incomplete data, followed by the generation of the ordering variable set $\Pi$ through $d$ generative iterations.
Upon transforming $\mathcal{G}^{\Pi}$ to $\mathcal{G}^{\Pi}_{g_{k}}$ as depicted in Figure \ref{fig:multiply}, we calculate rewards and subsequently update the actor and critic networks.
After sufficient iterations, the process yields $\mathcal{G}^*$, the optimal directed acyclic graph, with the highest reward for the manufacturing data.

\subparagraph{\textbf{States.}}
Preprocessing data using neural networks can enhance the ability to identify better orderings \cite{CORL}. 
Consequently, we define embedding $z^\Gamma$ generated by the incomplete and the complete data process as a state. 
The state space consists of embeddings from all variables, that is, $s_{t} = \{z_{1},\dots, z_{d}\}$. 
The initial state $s_{0}$ is obtained by concatenating all the embeddings without masking and using $s_{t}$ to represent the state at the $t$-th generative iteration of the variable selection. 
We need $d$ generative iterations to complete the ordering of variables; thus, the state transition in ($d-1$) times for each training process.

\subparagraph{\textbf{Actions.}}
We consider the selection of variables as action; thus, we define the action space as $A=\{x_{1},\dots,x_{d}\}$. 
Getting $\Pi$ necessitates $d$ generative iterations, and each generative iteration affects the next state.

\subparagraph{\textbf{State Transition.}}
The evolution of the state changes throughout the iterations of the selection of variables. 
In the $t$-th generative iteration, the decoder generates variable $x_{j}$ from $s_{t}$.
Then in a subsequent iteration $(t+1)$, we get the next state by masking the embedding $z_{j}$ from $s_{t}$, resulting in $s_{t+1} = Mask(s_{t},t)$ which is the same as the process in Equation (\ref{eq:masking}).

\subparagraph{\textbf{Reward and Optimization.}}
After getting $\mathcal{G}^{\Pi}_{g_k}$ by Section \ref{sec:ordering_graph_generating}, we utilize the Bayesian Information Criterion (BIC) score to evaluate the $\mathcal{G}^{\Pi}_{g_k}$. 
In causal graph construction, we determine a better graph that minimizes the BIC score which considers the noise variances to be equal \cite{GARL}; therefore, we define the reward function as $-BIC$. 
The reward function we use is :
\begin{equation}
\small
\begin{split}
R(g) &= -BIC(g) - D(g) \\
&=\log (\sum_{i = 1}^{d} (X^{r_{full}}_i - \hat X^{r_{full}}_i)) - \log(n) - |E(g)|\frac{\log (n)}{n} - D(g)
\end{split}
\end{equation}
where $n$ is the number of samples that are randomly chosen in each training iteration, and $|E(g)|$ is the number of edges in $g$. 
The penalty function $D(g)$ penalizes the absence of edges in the causal graph $g$, which must exist from expert knowledge.
We replace $g$ with $\mathcal{G}^{\Pi}_{g_k}$ for evaluating the graph from each training iteration. 
The optimization objective is to maximize the expected $J(\theta) = \mathbb{E}_{a\sim \pi_{\theta}(\cdot)} R$. 
In the $J(\theta)$, $\pi_{\theta}(\cdot)$ is the ordering graph generator model parameterized by $\theta$ which we use in Section \ref{sec:ordering_graph_generating} for the ordering graph generation process. 
We then update the actor for optimizing the policy gradient:
\begin{equation}
\nabla J(\theta) = \nabla \mathbb{E}_{\pi_{\theta}}\{\nabla_{\theta} \log \pi_{\theta}(\cdot)[R-b-V^{\pi_{\theta}}(s)]\}.
\label{eq:gradient}
\end{equation}
In Equation (\ref{eq:gradient}), the baseline $b$ is obtained by $b = \gamma \cdot b +(1 - \gamma)\cdot mean(R)$, where $\gamma$ is a discount rate to remove the bias of reward. 
$V^{\pi_{\theta}}(s)$ is the value of state $s$ when following a policy $\pi_{\theta}$ calculated by the critic network. 
\section{Experiments}
\begin{table*}
    \centering
    \small
    \caption{Overall performance evaluated by F1-score on synthetic data with different missing rates (MR) and the number of variables. The best result is highlighted in \textbf{Bold}, while the second best is \underline{underlined}. }
    \vspace{-10pt}
    \label{tab:syn_f1}
    \scalebox{1}{ %0.85
\renewcommand{\arraystretch}{0.8}
\newcommand{\specialcell}[2][c]{%
\begin{tabular}[#1]{@{}c@{}}#2\end{tabular}}

\begin{tabular}{cc|cccc|cccc|cccc}
    \toprule
    &\multicolumn{1}{c|}{Dataset}& \multicolumn{4}{c|}{$p50\_k20$} & \multicolumn{4}{c|}{$p100\_k30$} & \multicolumn{4}{c}{$p180\_k60$}\\ 
    \cmidrule{2-14}
    Model & MR &  50\% & 78\% & 82\% & 91\% & 50\% & 71\% & 83\% & 90\% & 50\% & 73\% & 83\% & 90\%  \\
    \midrule
    
    \multirow{2}{*}{Notears}  & Drop. & 0.4186 & 0.4186 & \underline{0.4186} & \underline{0.4186} & 0.2066 & 0.2066 & 0.2066 & 0.2066 & 0.0961 & 0.0961 & 0.0961 & 0.0961 \\
    
     & Imp. & 0.4808 & 0.4272 & 0.3761 & 0.3359 & 0.4135 & 0.3594 & 0.3043 & 0.2632 & 0.3405 & 0.3231 & 0.2810 & 0.2420\\
    
    \multirow{2}{*}{ICA-LiNGAM} & Drop. & 0.1351 & 0.1351 & 0.1351 & 0.1351 & - & - & - & - & - & - & - & -  \\
     & Imp. & 0.0833 & 0 & 0 & 0.0270 & 0.0671 & 0.0299 & 0.0286 & 0 & 0.0672 & 0.0453 & 0.0408 & 0.0310\\
    
    % \midrule
    % \midrule
    \multirow{2}{*}{GARL} & Drop. & 0.3553 & 0.3553 & 0.3553 & 0.3553 & 0.3553 & 0.3553 & 0.3553 & \underline{0.3553} & 0.3291 & 0.3291 & 0.3291 & \underline{0.3291}  \\
    & Imp. & 0.5258 & \underline{0.4510} & 0.4084 & 0.3860 & 0.4648 & 0.4444 & \underline{0.3955} & 0.2755 & 0.4256 & \underline{0.4590} & 0.3296 & 0.2960 \\
    
    \midrule
    MissDAG &  & \textbf{0.6880} & 0.3571 & 0.2524 & 0.2069 & \underline{0.5294} & \underline{0.5191} & 0.2186 & 0.1744 & \underline{0.4688} & 0.4397 & \underline{0.3362} & 0.2006 \\
    \midrule
    COKE &  & \underline{0.6241} & \textbf{0.6012} & \textbf{0.5976} & \textbf{0.6194} & \textbf{0.5350} & \textbf{0.5437} & \textbf{0.5338} & \textbf{0.5214} & \textbf{0.4821} & \textbf{0.4870} & \textbf{0.5214} & \textbf{0.5352} \\
    \bottomrule
\end{tabular}
    }
    \vspace{-4pt}
    
\end{table*}

% In Section \ref{sec:experiment setting}, we generate several synthetic datasets based on the recording methods of the manufacturing data and establish baselines.
% In Section \ref{sec:experiment setting},
% we describe the experimental settings, including the generation of synthetic datasets based on manufacturing data recording methods, the use of real-world data, baselines, and evaluation metrics.
% Section \ref{sec:result} presents results from both synthetic and real-world datasets, comparing our approach with benchmarks. 
% Section \ref{sec:ablation} discusses the impact of the components in our model.

\subsection{Experimental Setting}
\label{sec:experiment setting}
\subparagraph{\textbf{Synthetic Data.}}
% We generate data similar to manufacturing data, following the process used in \cite{HECSL} for semiconductor wafer data. 
% The data generation process is as same as the chronological order of manufacturing data, with directional influence between sensors across different machines, as shown in Figure \ref{fig:Motivation}. 
% For each variable, we generate data followed by \cite{lee2015learning}, without the discrete variables as we only focus on continuous data. 
% Consequently, data are sampled from a Gaussian graphical model:
We generate data similar to manufacturing data, following the process used in \cite{HECSL}  for semiconductor wafer data.
The data generation is the same as the chronological order of manufacturing, with directional influence between sensors across different machines, as shown in Figure \ref{fig:Motivation}. 
Data for each variable is generated following \cite{lee2015learning}, focusing on continuous data and sampled from a Gaussian graphical model: $X_i \sim \mathcal{N}({\mu}_i, {\Sigma}_i)$.
% In our setting, each variable has a different mean value but shares the same covariance matrix, therefore, averaging the data of its parent variables as its mean value.
% % We set the number of sensors $p$ as $\{50, 100, 180\}$ and the number of machines $k$ as $\{20, 30, 60\}$, and generate a sufficient number of samples $N=10000$ for each dataset. Conducting experiments on three datasets: $p50\_k20$, $p100\_k30$, and $p180\_k60$ to examine the effect on different numbers of variables in the corresponding missing rates. Also, we simulate expert knowledge by randomly selecting 10 edges as mandatory existing edges in datasets.
In our setting, each variable has a different mean value but shares the same covariance matrix, averaging the data of its parent variables as its mean. 
We set the number of sensors $p$ to $\{50, 100, 180\}$  and the number of machines $k$ to $\{20, 30, 60\}$, generating $N=10,000$ samples for each dataset. 
We conduct experiments on three datasets: $p50\_k20$, $p100\_k30$, and $p180\_k60$  to examine the effect of different numbers of variables on the corresponding missing rates. 
Additionally, we simulate expert knowledge by randomly selecting 10 edges as mandatory existing edges in the datasets.

% \subparagraph{\textbf{Missing Method.}}
To simulate product movement across machines according to specified recipes, we randomly select machines to represent scenarios where a product does not pass through a machine, resulting in missing values for all sensors in those selected machines.
Moreover, to reflect the ratios of complete data typically found in real-world datasets, we ensure that the entire dataset contains approximately 1\% of $X^{r_{full}}$ and 99\% of $X^{r_{miss}}$. 
With real-world data often showing missing proportions as high as 90\%, we establish a range of missing proportions for the three datasets, varying from 50\% to 90\%. 
Achieving precise missing proportions at the machine level is challenging. Therefore, we demonstrate four representative missing proportions for each dataset, such as 50\%, 73\%, 83\%, and 90\% in the $p180\_k60$.
\subparagraph{\textbf{Real-World Data.}}
Our real-world dataset records values following in Figure \ref{fig:Motivation}, for the 
 semiconductor wafer manufacturing process.
The dataset consists of 18 machines, 175 sensors, and 75 recipes. 
All values are continuous, with an average missing rate of 89.5\%.

\subparagraph{\textbf{Baselines.}}
Many existing causal discovery methods require complete datasets as input. 
Therefore, before applying these to incomplete datasets, we use two approaches: imputing missing values with MissForest \cite{missforest} and discarding samples with missing values, using only $X^{r_{full}}$.
In Table \ref{tab:syn_f1}, these methods are denoted as Imp. and Drop., respectively. 
Due to the unsuitability of constraint-based methods for high-dimensional datasets, we employ score-based methods such as Notears \cite{notears} and functional-based methods such as ICA-LiNGAM \cite{ICA-LiNGAM}. 
We compare the recent method GARL \cite{GARL}, incorporating expert knowledge and chronological order.
Since MissDAG \cite{missdag} and COKE are designed for causal discovery with missing data, we do not impute or drop samples with missing values, as they leverage incomplete and complete data. 
For baseline methods, we increase the number of training iterations to ensure convergence, keeping other parameters at their default settings.

\subparagraph{\textbf{Evaluation Metric.}}
In causal graph construction, excessive edge predictions yield higher recall scores but lower precision scores, whereas models predicting fewer edges exhibit better precision but lower recall. 
Both scenarios can lead to inefficiencies in manual edge selection for engineers. 
Therefore, we choose the harmonic mean of recall and precision, which is the F1-score, as the final evaluation metric for comparison of the baselines and COKE.
% ###

\subsection{Quantitative Results.}
\label{sec:result}
\begin{table}
    \centering
    \small
    \caption{Performance on a real-world dataset. }
    \vspace{-8pt}
    \label{tab:real_data}
    \scalebox{1}{ %
\renewcommand{\arraystretch}{0.8}
\newcommand{\specialcell}[2][c]{%
  \begin{tabular}[#1]{@{}c@{}}#2\end{tabular}}

\begin{tabular}{c|ccc}
    \toprule
    Model & Precision & Recall & F1 \\
    \midrule
    Notears & \underline{0.0822} & 0.0405 & \underline{0.0543}\\ 
    ICA-LiNGAM & 0.0111 & \underline{0.1284} & 0.0205\\
    GARL & 0.0124 & 0.0473 & 0.0194\\
    MissDAG & \textbf{0.2222} & 0.0135 & 0.0255  \\
    \midrule
    COKE & 0.0778 & \textbf{0.1419} & \textbf{0.1005} \\
    \bottomrule
\end{tabular}
    }
    \vspace{-13pt}
\end{table}
\begin{table*}
    \centering
    \small
    \caption{Model ablation F1-score performance on synthetic data with different MR and number of variables.}
    \vspace{-10pt}
    \label{tab:syn_abl_f1}
    \scalebox{1}{ %0.85
    % \renewcommand{\arraystretch}{1.2}
% \newcommand{\specialcell}[2][c]{%
%   \begin{tabular}[#1]{@{}c@{}}#2\end{tabular}}

% \begin{tabular}{c|cccc|cccc|cccc}
%     \toprule
%     & \multicolumn{4}{c|}{$p50\_k20$} & \multicolumn{4}{c|}{$p100\_k30$} & \multicolumn{4}{c}{$p180\_k60$}\\
%     \cmidrule{2-13}
%     Model & 50\% & 78\% & 82\% & 91\% & 50\% & 71\% & 83\% & 90\% & 50\% & 73\% & 83\% & 90\% \\
%     \midrule
%     w/o Incomplete data proc. & 0.5783 & 0.6016 & 0.5714 & 0.5912 & 0.5311 & 0.5094 & 0.5128 & 0.5046 & 0.4077 & 0.4661 & 0.4714 & 0.4545 \\
%     w/o knowledge graph& 0.4571 & 0.4162 & 0.4424 & 0.4558 & 0.3853 & 0.3721 & 0.3922 & 0.3821 & 0.3864 & 0.3623 & 0.3680 & 0.3520 \\
%     \midrule
%     Proposed & \textbf{0.5926} & \textbf{0.6240} & \textbf{0.5976} & \textbf{0.5963} & \textbf{0.5443} & \textbf{0.5461} & \textbf{0.5561} & \textbf{0.5110} & \textbf{0.4928} & \textbf{0.4726} & \textbf{0.4809} & \textbf{0.4850}\\
    
%     \bottomrule
% \end{tabular}

% \xmark
% \cmark

% \renewcommand{\arraystretch}{1.2}
\renewcommand{\arraystretch}{0.8}
\newcommand{\specialcell}[2][c]{%
  \begin{tabular}[#1]{@{}c@{}}#2\end{tabular}}

\begin{tabular}{ccc|cccc|cccc|cccc}
    \toprule
    \multicolumn{3}{c|}{}& \multicolumn{4}{c|}{$p50\_k20$} & \multicolumn{4}{c|}{$p100\_k30$} & \multicolumn{4}{c}{$p180\_k60$}\\
    \cmidrule{4-15}
    CO & EK & Incomp. & 50\% & 78\% & 82\% & 91\% & 50\% & 71\% & 83\% & 90\% & 50\% & 73\% & 83\% & 90\% \\
    \midrule
    % % ----f-m
    % & & \ding{51} & 0.4519 & 0.4364 & 0.4424 & 0.4498 & 0.3733 & 0.3700 & 0.3688 & 0.3679 &  0.3706 & 0.3741 & 0.3806 & 0.3660 \\
    % --d-f-m
    &\ding{51}&\ding{51} & 0.4293 & 0.4372& 0.4579& 0.4585& 0.3750& 0.3750& 0.3814 &0.3577 & 0.3741 & 0.3647 & 0.3869 &  0.3717\\	
    % t---f-m
    \ding{51} &  & \ding{51} & \underline{0.5972} & 0.5714 & \underline{0.5818} & \underline{0.5972} & \underline{0.5016} & \underline{0.5321} & 0.5204 & 0.5118 & \underline{0.4717} & \underline{0.4809} & \underline{0.5143} &  \underline{0.5297} \\
    % t-d-f--
    \ding{51} & \ding{51}  & & 0.5960 & \underline{0.5976} & 0.5765 & 0.5935 & \underline{0.5016} & 0.5163 & \underline{0.5243} & \underline{0.5152} & 0.4455 & 0.4708 & 0.4573 &  0.4750\\
    \midrule
    % t-d-f-m
    \ding{51} & \ding{51} & \ding{51} & \textbf{0.6241} & \textbf{0.6012} & \textbf{0.5976} & \textbf{0.6194} & \textbf{0.5350} & \textbf{0.5437} & \textbf{0.5338} & \textbf{0.5214} & \textbf{0.4821} & \textbf{0.4870} & \textbf{0.5214}	& \textbf{0.5352}

    \\
    
    \bottomrule
\end{tabular}
    \vspace{-4pt}
    }
\end{table*}
\subparagraph{\textbf{Results for Synthetic Data.}}
We evaluated our approach across three datasets: $p50\_k20$, $p100\_k30$, and $p180\_k60$. 
Across each dataset, we examined four missing rate settings from approximately 50\% to 90\%. 
In total, our evaluation covered twelve datasets, with results presented in Table \ref{tab:syn_f1}.
All the experiments were conducted on an Intel Core i7 12700 with an NVIDIA RTX 3060.
We made the following observations:
% Comparison on Missing rate
% 1) COKE exhibits low sensitivity to increasing missingness. 
% While the F1-score of MissDAG notably declines with rising missing proportions from 50\% to 90\% across variable counts of 50, 100, and 180, with reductions of 69.9\%, 67.4\%, and 57.2\%, respectively, COKE exhibits much lower sensitivity, with decreases of only 0.7\%, 2.5\%, and -11.0\% respectively. 
1) COKE exhibits low sensitivity to increasing missingness.
We compared the F1-score decrement with missing ratios of 50\% and 90\% for variable counts of $p50$, $p100$, and $p180$. 
MissDAG showed decreases of 69.9\%, 67.4\%, and 57.2\%, and GARL(Imp.) showed 26.5\%, 40.7\%,  and 30.4\%, respectively.  
COKE showed minimal performance drops of 0.7\%, 2.5\%, and -11.0\%. 
In contrast, MissDAG and GARL experienced significantly greater declines when increasing missing ratios.
% While COKE exhibits only 0.7\%, 2.5\%, and -11.0\% decreases. 
% The performance drop in MissDAG across different variable quantities is significantly greater than that of COKE. 
% This experiment demonstrates COKE's effective use of recipes, ensuring that an increase in the proportion of missing data does not affect causal discovery.
% Comparison of certain method ex: GARL and Missdag
2) COKE had a significant advantage in handling high proportions of missing values. 
% We evaluated the performance of GARL(Imp.) and MissDAG under missing rates of 50\%, 75\%, 83\%, and 90\%. 
% When comparing COKE to GARL, the average F1-score improvements were 15.6\%, 20.5\%, 46.4\%, and 76.8\%, respectively. 
% In comparison with MissDAG, the improvements of using COKE were -1.7\%, 27.9\%, 66.5\%, and 186.5\%, respectively. 
% Compared to GARL and MissDAG, the enhancement in performance for high missing rates far exceeds that for low missing rates. This demonstrates that incomplete data processing in COKE effectively leverages missing values to construct causal graphs. Therefore, COKE is suitable for scenarios with high proportions of missing values.
We evaluated GARL(Imp.) and MissDAG under missing rates of 50\%, 75\%, 83\%, and 90\%. 
Compared to GARL, COKE exhibited average F1-score improvements of 15.6\%, 20.5\%, 46.4\%, and 76.8\%, respectively. Similarly, compared to MissDAG, COKE showed improvements of -1.7\%, 27.9\%, 66.5\%, and 189.5\%, respectively.  
COKE's performance enhancement was significantly higher for high missing rates compared to low missing rates, indicating its effectiveness in leveraging incomplete data for constructing causal graphs by incomplete data process. 
Hence, COKE is suitable for scenarios with high proportions of missing values.
% Comparison of the number of variables
3) Leveraging expert knowledge and chronological order with RL-based methods is insensitive to increasing variable numbers. 
% Unlike Notears(Imp.) and MissDAG, which experience significant drops in f1-score (13.9\% and 23.0\%, respectively) with variable count increases from 50 to 100, and even more drops (29.1\%, and 31.8\%, respectively) with variable count increases from 50 to 180, COKE and GARL(Imp.) training with RL architecture demonstrate the robustness. 
Notears(Imp.) and MissDAG experienced significant F1-score drops as the variable count increased: 13.9\% and 23.0\% when increasing from $p50$ to $p100$, and 29.1\% and 31.8\% when increasing from $p50$ to $p180$.
In contrast, as the variable count increased from $p50$ to $p100$ and $p50$ to $p180$, the F1-score decreased for COKE and GARL(Imp.), averaging 12.9\% and 20.8\%, respectively.
COKE and GARL use expert knowledge and chronological order within an RL architecture, demonstrating robust performance even as the number of variables increases.
% ###
% Imputation vs Drop
4) The impact of inductive bias introduced by imputation was reduced by leveraging expert knowledge and chronological order. 
Some experiments showed that dropping missing values yields better results than imputation. For instance, Notears(Drop) outperformed Notears(Imp) in the $p50\_k20$ dataset with 90\% missing values. 
Compared to GARL, which incorporates expert knowledge and chronological order, Notears and ICA-LiNGAM exhibited significant differences with or without imputation, highlighting the importance of reducing inductive bias by incorporating information. 

% Some experiments indicate that dropping missing values yields better results than imputation.
% For instance, Notears (Drop) outperforms Notears (Imp) in scenarios like the $p50\_k20$ dataset with 90\% 
%  of missing values.
% Compared to GARL, which incorporates expert knowledge and chronological order, Notears and ICA-LiNGAM exhibit significant differences with or without imputation.
% The incorporation of information can effectively reduce the impact of inductive bias.
ICA-LiNGAM struggles with synthetic data under varied generation settings, leading to an F1-score of zero. 
It also requires more samples than variables and fails in settings like $p100$ and $p180$ with dropped missing values. 
Different missing proportions at the same variable count were generated by introducing missing values while keeping a small proportion of complete data at 1\%. 
Consequently, dataset variations with different missing ratios resulted in similar datasets. 
In experiments using Notears(Drop.), ICA-LiNGAM(Drop.), and GARL(Drop.), F1-score differences were attributed solely to variations in the variable count across datasets.

\subparagraph{\textbf{Results for Real-World Data.}}
% Our real-world dataset records the sequential values between machines in the wafer manufacturing process, comprising 18 machines, 175 sensors, and 75 recipes. 
% All values are continuous, with a missing rate of 89.5\%. 
Assuming nonlinear relationships between variables, Table \ref{tab:real_data} indicates that COKE has the best performance on the F1-score. 
% that our proposed method may not perform as expected in precision. 
Our method may not meet precision expectations as MissDAG predicts only nine edges, while our model predicts hundreds, resulting in lower precision. 
Predicting more edges in manufacturing dataset graph construction helps alleviate engineers' workload. 
Thus, despite MissDAG's superior precision, it is impractical for real-world use.

\begin{figure}[h]
    \vspace{-5pt}
     \small
     \centering
     \begin{subfigure}[b]{0.22\textwidth}
        \small
         \centering
         \includegraphics[width=\linewidth]{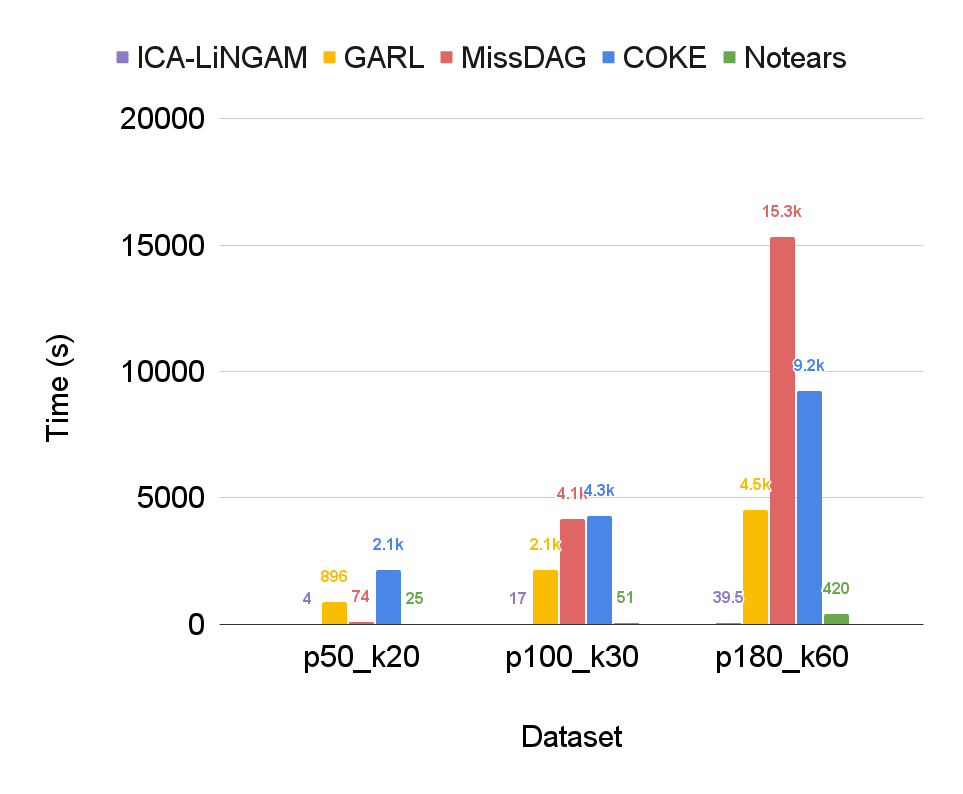}
         \vspace{-20pt}
         \caption{Time complexity.}
     \end{subfigure}
     \hfill
     \begin{subfigure}[b]{0.22\textwidth}
         \small
         \centering
         \includegraphics[width=\linewidth]{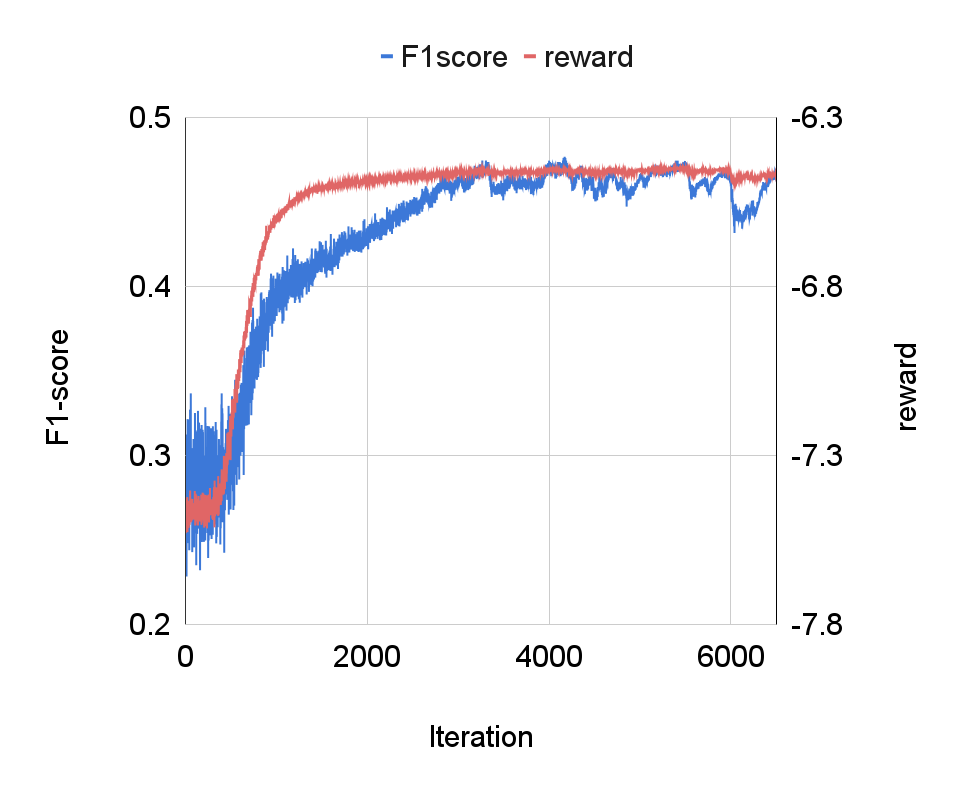}
         \vspace{-20pt}
         \caption{Training process.}
     \end{subfigure}
        \vspace{-10pt}
        \caption{Training information in COKE and other baselines.}
        \label{fig:case-study}
    \vspace{-10pt}
\end{figure}

\subparagraph{\textbf{Execution Time.}}
Figure \ref{fig:case-study} (a) shows the training times for each model.
MissDAG exhibited a longer training time for numerous variables, indicating inefficiency in high-dimensional causal discovery.
In contrast, GARL's training time increased less with more variables, demonstrating the efficiency of its RL-based architecture.
While COKE requires more time in datasets with fewer variables, it significantly reduces processing time compared to MissDAG in datasets with more variables. 
COKE benefits from RL-based techniques, demonstrating consistent time efficiency up to $p180$.

\subparagraph{\textbf{Visualizations of the Optimization Process.}}
% Figure \ref{fig:case-study} (b) illustrates the iteration status between the F1-score and reward during the training process of COKE.
% It can be observed that even including incomplete data processing, our model still manages to steadily improve the F1-score and converge.
The designed reward function proves beneficial for causal graph construction, as evidenced by the positive correlation between the F1 score and the reward. 
In Figure \ref{fig:case-study} (b), the iteration status between the F1-score and reward during COKE's training process illustrates this correlation. 
Despite incorporating an incomplete data process, our model consistently enhances the F1-score and achieves convergence.

\subsection{Ablation Study}
\label{sec:ablation}
To evaluate the importance of chronological order (CO), expert knowledge (EK), and the incomplete data process (Incomp.) in our model, we conducted experiments in Table \ref{tab:syn_abl_f1}.
Chronological ordering information is crucial for accurate causal graph construction, as indicated by the significant performance drop observed when this aspect is neglected.
This results in inferior performance across all 12 settings, failing to achieve even the second-best result in any of them.
The impact of the incomplete data process is significant as the number of variables increases.
Specifically, results show a marked decrease in performance when the incomplete data process is not applied to datasets with $p180$ variables.
These experiments show the essential roles that chronological order, expert knowledge, and the incomplete data process play in our model, demonstrating their importance in achieving robust results.

\section{Conclusion}
In this paper, we introduce COKE, a method designed to construct causal graphs by leveraging expert knowledge and chronological order in datasets with high missing values. 
COKE utilizes manufacturing recipes to maximize the utility of samples with missing values, enabling graph construction without data imputation. 
By employing the graph attention network, COKE integrates expert knowledge and chronological order into the actor-critic training process. 
It utilizes the BIC score as a reward function to optimize the graph generation model, eliciting the final causal graph with the maximum reward.
Experimental results show that in synthetic data with multiple representative missing proportions and variable counts, COKE outperformed several strong baselines across various settings showcasing the effectiveness of our framework.
% Thus, we believe that COKE could effectively construct causal graphs not only in manufacturing datasets but also in other domains like reservoir systems, where expert knowledge and chronological order are essential for establishing causal relationships.
We believe COKE could flexibly construct causal graphs not only in manufacturing datasets but also in other domains where leveraging knowledge is crucial for establishing accurate causal relationships.
% In future work, we aim to further validate the applicability of COKE across diverse domains such as reservoir systems, and explore its potential in addressing specific challenges unique to each domain.
In the future, how to build causal graphs using only samples with missing values will be further investigated within this framework, as real-world data often lack complete datasets.

\bibliographystyle{ACM-Reference-Format}
\balance
\bibliography{reference}


\begin{thebibliography}{31}

%%% the Web address.

\ifx \showCODEN    \undefined \def \showCODEN     #1{\unskip}     \fi
\ifx \showDOI      \undefined \def \showDOI       #1{#1}\fi
\ifx \showISBNx    \undefined \def \showISBNx     #1{\unskip}     \fi
\ifx \showISBNxiii \undefined \def \showISBNxiii  #1{\unskip}     \fi
\ifx \showISSN     \undefined \def \showISSN      #1{\unskip}     \fi
\ifx \showLCCN     \undefined \def \showLCCN      #1{\unskip}     \fi
\ifx \shownote     \undefined \def \shownote      #1{#1}          \fi
\ifx \showarticletitle \undefined \def \showarticletitle #1{#1}   \fi
\ifx \showURL      \undefined \def \showURL       {\relax}        \fi
% invisible to TeX
\providecommand\bibfield[2]{#2}
\providecommand\bibinfo[2]{#2}
\providecommand\natexlab[1]{#1}
\providecommand\showeprint[2][]{arXiv:#2}

\bibitem[Abu-Samah et~al\mbox{.}(2015)]%
        {cdmf-failure}
\bibfield{author}{\bibinfo{person}{A Abu-Samah}, \bibinfo{person}{MK Shahzad}, \bibinfo{person}{E Zamai}, {and} \bibinfo{person}{A~Ben Said}.} \bibinfo{year}{2015}\natexlab{}.
\newblock \showarticletitle{Failure prediction methodology for improved proactive maintenance using Bayesian approach}.
\newblock \bibinfo{journal}{\emph{IFAC-PapersOnLine}} \bibinfo{volume}{48}, \bibinfo{number}{21} (\bibinfo{year}{2015}), \bibinfo{pages}{844--851}.
\newblock


\bibitem[B{\"{u}}hlmann et~al\mbox{.}(2013)]%
        {CAM}
\bibfield{author}{\bibinfo{person}{Peter B{\"{u}}hlmann}, \bibinfo{person}{Jonas Peters}, {and} \bibinfo{person}{Jan Ernest}.} \bibinfo{year}{2013}\natexlab{}.
\newblock \showarticletitle{{CAM:} Causal Additive Models, high-dimensional order search and penalized regression}.
\newblock \bibinfo{journal}{\emph{CoRR}}  \bibinfo{volume}{abs/1310.1533} (\bibinfo{year}{2013}).
\newblock


\bibitem[Clavijo et~al\mbox{.}(2021)]%
        {cdmf-faultdetect}
\bibfield{author}{\bibinfo{person}{Nayher Clavijo}, \bibinfo{person}{Afr{\^a}nio Melo}, \bibinfo{person}{Rafael~M Soares}, \bibinfo{person}{Luiz Felipe de~O Campos}, \bibinfo{person}{Tiago Lemos}, \bibinfo{person}{Maur{\'\i}cio~M C{\^a}mara}, \bibinfo{person}{Thiago~K Anzai}, \bibinfo{person}{Fabio~C Diehl}, \bibinfo{person}{Pedro~H Thompson}, {and} \bibinfo{person}{Jos{\'e}~Carlos Pinto}.} \bibinfo{year}{2021}\natexlab{}.
\newblock \showarticletitle{Variable selection for fault detection based on causal discovery methods: Analysis of an actual industrial case}.
\newblock \bibinfo{journal}{\emph{Processes}} \bibinfo{volume}{9}, \bibinfo{number}{3} (\bibinfo{year}{2021}), \bibinfo{pages}{544}.
\newblock


\bibitem[Colledani et~al\mbox{.}(2014)]%
        {mf-auto}
\bibfield{author}{\bibinfo{person}{Marcello Colledani}, \bibinfo{person}{Tullio Tolio}, \bibinfo{person}{Anath Fischer}, \bibinfo{person}{Benoit Iung}, \bibinfo{person}{Gisela Lanza}, \bibinfo{person}{Robert Schmitt}, {and} \bibinfo{person}{J{\'o}zsef V{\'a}ncza}.} \bibinfo{year}{2014}\natexlab{}.
\newblock \showarticletitle{Design and management of manufacturing systems for production quality}.
\newblock \bibinfo{journal}{\emph{Cirp Annals}} \bibinfo{volume}{63}, \bibinfo{number}{2} (\bibinfo{year}{2014}), \bibinfo{pages}{773--796}.
\newblock


\bibitem[Ezukwoke et~al\mbox{.}(2024)]%
        {mf-imp1}
\bibfield{author}{\bibinfo{person}{Kenneth Ezukwoke}, \bibinfo{person}{Anis Hoayek}, \bibinfo{person}{Mireille Batton-Hubert}, \bibinfo{person}{Xavier Boucher}, \bibinfo{person}{Pascal Gounet}, {and} \bibinfo{person}{J{\'e}r{\^o}me Adrian}.} \bibinfo{year}{2024}\natexlab{}.
\newblock \showarticletitle{Big GCVAE: decision-making with adaptive transformer model for failure root cause analysis in semiconductor industry}.
\newblock \bibinfo{journal}{\emph{Journal of Intelligent Manufacturing}} (\bibinfo{year}{2024}), \bibinfo{pages}{1--16}.
\newblock


\bibitem[Gao et~al\mbox{.}(2022)]%
        {missdag}
\bibfield{author}{\bibinfo{person}{Erdun Gao}, \bibinfo{person}{Ignavier Ng}, \bibinfo{person}{Mingming Gong}, \bibinfo{person}{Li Shen}, \bibinfo{person}{Wei Huang}, \bibinfo{person}{Tongliang Liu}, \bibinfo{person}{Kun Zhang}, {and} \bibinfo{person}{Howard~D. Bondell}.} \bibinfo{year}{2022}\natexlab{}.
\newblock \showarticletitle{MissDAG: Causal Discovery in the Presence of Missing Data with Continuous Additive Noise Models}. In \bibinfo{booktitle}{\emph{NeurIPS}}.
\newblock


\bibitem[Gao and Cai(2023)]%
        {MICwgan}
\bibfield{author}{\bibinfo{person}{Yanyang Gao} {and} \bibinfo{person}{Qingsong Cai}.} \bibinfo{year}{2023}\natexlab{}.
\newblock \showarticletitle{A WGAN-based Missing Data Causal Discovery Method}. In \bibinfo{booktitle}{\emph{2023 4th International Conference on Big Data, Artificial Intelligence and Internet of Things Engineering (ICBAIE)}}. IEEE, \bibinfo{pages}{136--139}.
\newblock


\bibitem[Gharahbagheri et~al\mbox{.}(2015)]%
        {mf-fluid}
\bibfield{author}{\bibinfo{person}{H Gharahbagheri}, \bibinfo{person}{S Imtiaz}, \bibinfo{person}{Faisal Khan}, {and} \bibinfo{person}{S Ahmed}.} \bibinfo{year}{2015}\natexlab{}.
\newblock \showarticletitle{Causality analysis for root cause diagnosis in Fluid Catalytic Cracking unit}.
\newblock \bibinfo{journal}{\emph{IFAC-PapersOnLine}} \bibinfo{volume}{48}, \bibinfo{number}{21} (\bibinfo{year}{2015}), \bibinfo{pages}{838--843}.
\newblock


\bibitem[Hagedorn et~al\mbox{.}(2022)]%
        {cd-mfp}
\bibfield{author}{\bibinfo{person}{Christopher Hagedorn}, \bibinfo{person}{Johannes Huegle}, {and} \bibinfo{person}{Rainer Schlosser}.} \bibinfo{year}{2022}\natexlab{}.
\newblock \showarticletitle{Understanding unforeseen production downtimes in manufacturing processes using log data-driven causal reasoning}.
\newblock \bibinfo{journal}{\emph{J. Intell. Manuf.}} \bibinfo{volume}{33}, \bibinfo{number}{7} (\bibinfo{year}{2022}), \bibinfo{pages}{2027--2043}.
\newblock


\bibitem[Huang et~al\mbox{.}(2020)]%
        {cd-rl-miss}
\bibfield{author}{\bibinfo{person}{Xiaoshui Huang}, \bibinfo{person}{Fujin Zhu}, \bibinfo{person}{Lois Holloway}, {and} \bibinfo{person}{Ali Haidar}.} \bibinfo{year}{2020}\natexlab{}.
\newblock \showarticletitle{Causal discovery from incomplete data using an encoder and reinforcement learning}.
\newblock \bibinfo{journal}{\emph{arXiv preprint arXiv:2006.05554}} (\bibinfo{year}{2020}).
\newblock


\bibitem[Huegle et~al\mbox{.}(2020)]%
        {cd-app-car}
\bibfield{author}{\bibinfo{person}{Johannes Huegle}, \bibinfo{person}{Christopher Hagedorn}, {and} \bibinfo{person}{Matthias Uflacker}.} \bibinfo{year}{2020}\natexlab{}.
\newblock \showarticletitle{How Causal Structural Knowledge Adds Decision-Support in Monitoring of Automotive Body Shop Assembly Lines}. In \bibinfo{booktitle}{\emph{{IJCAI}}}. \bibinfo{publisher}{ijcai.org}, \bibinfo{pages}{5246--5248}.
\newblock


\bibitem[Ikram et~al\mbox{.}(2022)]%
        {mf-rca1}
\bibfield{author}{\bibinfo{person}{Azam Ikram}, \bibinfo{person}{Sarthak Chakraborty}, \bibinfo{person}{Subrata Mitra}, \bibinfo{person}{Shiv~Kumar Saini}, \bibinfo{person}{Saurabh Bagchi}, {and} \bibinfo{person}{Murat Kocaoglu}.} \bibinfo{year}{2022}\natexlab{}.
\newblock \showarticletitle{Root Cause Analysis of Failures in Microservices through Causal Discovery}. In \bibinfo{booktitle}{\emph{NeurIPS}}.
\newblock


\bibitem[Kwak and Kim(2012)]%
        {recipe2miss_1}
\bibfield{author}{\bibinfo{person}{Doh{-}Soon Kwak} {and} \bibinfo{person}{Kwang{-}Jae Kim}.} \bibinfo{year}{2012}\natexlab{}.
\newblock \showarticletitle{A data mining approach considering missing values for the optimization of semiconductor-manufacturing processes}.
\newblock \bibinfo{journal}{\emph{Expert Syst. Appl.}} \bibinfo{volume}{39}, \bibinfo{number}{3} (\bibinfo{year}{2012}), \bibinfo{pages}{2590--2596}.
\newblock


\bibitem[Le et~al\mbox{.}(2019)]%
        {fpc}
\bibfield{author}{\bibinfo{person}{Thuc~Duy Le}, \bibinfo{person}{Tao Hoang}, \bibinfo{person}{Jiuyong Li}, \bibinfo{person}{Lin Liu}, \bibinfo{person}{Huawen Liu}, {and} \bibinfo{person}{Shu Hu}.} \bibinfo{year}{2019}\natexlab{}.
\newblock \showarticletitle{A Fast {PC} Algorithm for High Dimensional Causal Discovery with Multi-Core PCs}.
\newblock \bibinfo{journal}{\emph{{IEEE} {ACM} Trans. Comput. Biol. Bioinform.}} \bibinfo{volume}{16}, \bibinfo{number}{5} (\bibinfo{year}{2019}), \bibinfo{pages}{1483--1495}.
\newblock


\bibitem[Lee and Hastie(2015)]%
        {lee2015learning}
\bibfield{author}{\bibinfo{person}{Jason~D Lee} {and} \bibinfo{person}{Trevor~J Hastie}.} \bibinfo{year}{2015}\natexlab{}.
\newblock \showarticletitle{Learning the structure of mixed graphical models}.
\newblock \bibinfo{journal}{\emph{Journal of Computational and Graphical Statistics}} \bibinfo{volume}{24}, \bibinfo{number}{1} (\bibinfo{year}{2015}), \bibinfo{pages}{230--253}.
\newblock


\bibitem[Liang et~al\mbox{.}(2004)]%
        {mf-complexity}
\bibfield{author}{\bibinfo{person}{Steven~Y Liang}, \bibinfo{person}{Rogelio~L Hecker}, {and} \bibinfo{person}{Robert~G Landers}.} \bibinfo{year}{2004}\natexlab{}.
\newblock \showarticletitle{Machining process monitoring and control: the state-of-the-art}.
\newblock \bibinfo{journal}{\emph{J. Manuf. Sci. Eng.}} \bibinfo{volume}{126}, \bibinfo{number}{2} (\bibinfo{year}{2004}), \bibinfo{pages}{297--310}.
\newblock


\bibitem[Marazopoulou et~al\mbox{.}(2016)]%
        {cdmf-md}
\bibfield{author}{\bibinfo{person}{Katerina Marazopoulou}, \bibinfo{person}{Rumi Ghosh}, \bibinfo{person}{Prasanth Lade}, {and} \bibinfo{person}{David~D. Jensen}.} \bibinfo{year}{2016}\natexlab{}.
\newblock \showarticletitle{Causal Discovery for Manufacturing Domains}.
\newblock \bibinfo{journal}{\emph{CoRR}}  \bibinfo{volume}{abs/1605.04056} (\bibinfo{year}{2016}).
\newblock


\bibitem[Morales{-}Alvarez et~al\mbox{.}(2022)]%
        {VISL}
\bibfield{author}{\bibinfo{person}{Pablo Morales{-}Alvarez}, \bibinfo{person}{Wenbo Gong}, \bibinfo{person}{Angus Lamb}, \bibinfo{person}{Simon Woodhead}, \bibinfo{person}{Simon~Peyton Jones}, \bibinfo{person}{Nick Pawlowski}, \bibinfo{person}{Miltiadis Allamanis}, {and} \bibinfo{person}{Cheng Zhang}.} \bibinfo{year}{2022}\natexlab{}.
\newblock \showarticletitle{Simultaneous Missing Value Imputation and Structure Learning with Groups}. In \bibinfo{booktitle}{\emph{NeurIPS}}.
\newblock


\bibitem[Qiao et~al\mbox{.}(2024)]%
        {cd-miss2}
\bibfield{author}{\bibinfo{person}{Jie Qiao}, \bibinfo{person}{Zhengming Chen}, \bibinfo{person}{Jianhua Yu}, \bibinfo{person}{Ruichu Cai}, {and} \bibinfo{person}{Zhifeng Hao}.} \bibinfo{year}{2024}\natexlab{}.
\newblock \showarticletitle{Identification of Causal Structure in the Presence of Missing Data with Additive Noise Model}. In \bibinfo{booktitle}{\emph{{AAAI}}}. \bibinfo{pages}{20516--20523}.
\newblock


\bibitem[Qin et~al\mbox{.}(2022)]%
        {mf-rca2}
\bibfield{author}{\bibinfo{person}{Kai Qin}, \bibinfo{person}{Lei Chen}, \bibinfo{person}{Jintao Shi}, \bibinfo{person}{Zhenxing Li}, {and} \bibinfo{person}{Kuangrong Hao}.} \bibinfo{year}{2022}\natexlab{}.
\newblock \showarticletitle{Root cause analysis of industrial faults based on binary extreme gradient boosting and temporal causal discovery network}.
\newblock \bibinfo{journal}{\emph{Chemometrics and Intelligent Laboratory Systems}}  \bibinfo{volume}{225} (\bibinfo{year}{2022}), \bibinfo{pages}{104559}.
\newblock


\bibitem[Rubin(1976)]%
        {IMS}
\bibfield{author}{\bibinfo{person}{Donald~B Rubin}.} \bibinfo{year}{1976}\natexlab{}.
\newblock \showarticletitle{Inference and missing data}.
\newblock \bibinfo{journal}{\emph{Biometrika}} \bibinfo{volume}{63}, \bibinfo{number}{3} (\bibinfo{year}{1976}), \bibinfo{pages}{581--592}.
\newblock


\bibitem[Shimizu et~al\mbox{.}(2006)]%
        {ICA-LiNGAM}
\bibfield{author}{\bibinfo{person}{Shohei Shimizu}, \bibinfo{person}{Patrik~O. Hoyer}, \bibinfo{person}{Aapo Hyv{\"{a}}rinen}, {and} \bibinfo{person}{Antti~J. Kerminen}.} \bibinfo{year}{2006}\natexlab{}.
\newblock \showarticletitle{A Linear Non-Gaussian Acyclic Model for Causal Discovery}.
\newblock \bibinfo{journal}{\emph{J. Mach. Learn. Res.}}  \bibinfo{volume}{7} (\bibinfo{year}{2006}), \bibinfo{pages}{2003--2030}.
\newblock


\bibitem[Sim et~al\mbox{.}(2014)]%
        {cdmf-dm}
\bibfield{author}{\bibinfo{person}{Hyunsik Sim}, \bibinfo{person}{Doowon Choi}, {and} \bibinfo{person}{Chang~Ouk Kim}.} \bibinfo{year}{2014}\natexlab{}.
\newblock \showarticletitle{A data mining approach to the causal analysis of product faults in multi-stage PCB manufacturing}.
\newblock \bibinfo{journal}{\emph{International journal of precision engineering and manufacturing}}  \bibinfo{volume}{15} (\bibinfo{year}{2014}), \bibinfo{pages}{1563--1573}.
\newblock


\bibitem[Stekhoven and B{\"{u}}hlmann(2012)]%
        {missforest}
\bibfield{author}{\bibinfo{person}{Daniel~J. Stekhoven} {and} \bibinfo{person}{Peter B{\"{u}}hlmann}.} \bibinfo{year}{2012}\natexlab{}.
\newblock \showarticletitle{MissForest - non-parametric missing value imputation for mixed-type data}.
\newblock \bibinfo{journal}{\emph{Bioinform.}} \bibinfo{volume}{28}, \bibinfo{number}{1} (\bibinfo{year}{2012}), \bibinfo{pages}{112--118}.
\newblock


\bibitem[Teyssier and Koller(2005)]%
        {Order-BS}
\bibfield{author}{\bibinfo{person}{Marc Teyssier} {and} \bibinfo{person}{Daphne Koller}.} \bibinfo{year}{2005}\natexlab{}.
\newblock \showarticletitle{Ordering-Based Search: {A} Simple and Effective Algorithm for Learning Bayesian Networks}. In \bibinfo{booktitle}{\emph{{UAI} '05, Proceedings of the 21st Conference in Uncertainty in Artificial Intelligence, Edinburgh, Scotland, July 26-29, 2005}}. \bibinfo{publisher}{{AUAI} Press}, \bibinfo{pages}{548--549}.
\newblock


\bibitem[Tu et~al\mbox{.}(2018)]%
        {mvpc}
\bibfield{author}{\bibinfo{person}{Ruibo Tu}, \bibinfo{person}{Cheng Zhang}, \bibinfo{person}{Paul Ackermann}, \bibinfo{person}{Hedvig Kjellstr{\"{o}}m}, {and} \bibinfo{person}{Kun Zhang}.} \bibinfo{year}{2018}\natexlab{}.
\newblock \showarticletitle{Causal discovery in the presence of missing data}.
\newblock \bibinfo{journal}{\emph{CoRR}}  \bibinfo{volume}{abs/1807.04010} (\bibinfo{year}{2018}).
\newblock


\bibitem[Wang et~al\mbox{.}(2021)]%
        {CORL}
\bibfield{author}{\bibinfo{person}{Xiaoqiang Wang}, \bibinfo{person}{Yali Du}, \bibinfo{person}{Shengyu Zhu}, \bibinfo{person}{Liangjun Ke}, \bibinfo{person}{Zhitang Chen}, \bibinfo{person}{Jianye Hao}, {and} \bibinfo{person}{Jun Wang}.} \bibinfo{year}{2021}\natexlab{}.
\newblock \showarticletitle{Ordering-Based Causal Discovery with Reinforcement Learning}, \bibfield{editor}{\bibinfo{person}{Zhi{-}Hua Zhou}} (Ed.). \bibinfo{publisher}{ijcai.org}, \bibinfo{pages}{3566--3573}.
\newblock


\bibitem[Wang et~al\mbox{.}(2020)]%
        {ICL}
\bibfield{author}{\bibinfo{person}{Yuhao Wang}, \bibinfo{person}{Vlado Menkovski}, \bibinfo{person}{Hao Wang}, \bibinfo{person}{Xin Du}, {and} \bibinfo{person}{Mykola Pechenizkiy}.} \bibinfo{year}{2020}\natexlab{}.
\newblock \showarticletitle{Causal Discovery from Incomplete Data: {A} Deep Learning Approach}.
\newblock \bibinfo{journal}{\emph{CoRR}}  \bibinfo{volume}{abs/2001.05343} (\bibinfo{year}{2020}).
\newblock


\bibitem[Yang et~al\mbox{.}(2023b)]%
        {GARL}
\bibfield{author}{\bibinfo{person}{Dezhi Yang}, \bibinfo{person}{Guoxian Yu}, \bibinfo{person}{Jun Wang}, \bibinfo{person}{Zhongmin Yan}, {and} \bibinfo{person}{Maozu Guo}.} \bibinfo{year}{2023}\natexlab{b}.
\newblock \showarticletitle{Causal Discovery by Graph Attention Reinforcement Learning}. In \bibinfo{booktitle}{\emph{Proceedings of the 2023 {SIAM} International Conference on Data Mining, {SDM} 2023, Minneapolis-St. Paul Twin Cities, MN, USA, April 27-29, 2023}}. \bibinfo{publisher}{{SIAM}}, \bibinfo{pages}{28--36}.
\newblock


\bibitem[Yang et~al\mbox{.}(2023a)]%
        {HECSL}
\bibfield{author}{\bibinfo{person}{Yu Yang}, \bibinfo{person}{Sthitie Bom}, {and} \bibinfo{person}{Xiaotong Shen}.} \bibinfo{year}{2023}\natexlab{a}.
\newblock \showarticletitle{A hierarchical ensemble causal structure learning approach for wafer manufacturing}.
\newblock \bibinfo{journal}{\emph{Journal of Intelligent Manufacturing}} (\bibinfo{year}{2023}), \bibinfo{pages}{1--18}.
\newblock


\bibitem[Zheng et~al\mbox{.}(2018)]%
        {notears}
\bibfield{author}{\bibinfo{person}{Xun Zheng}, \bibinfo{person}{Bryon Aragam}, \bibinfo{person}{Pradeep Ravikumar}, {and} \bibinfo{person}{Eric~P. Xing}.} \bibinfo{year}{2018}\natexlab{}.
\newblock \showarticletitle{DAGs with {NO} {TEARS:} Continuous Optimization for Structure Learning}. In \bibinfo{booktitle}{\emph{NeurIPS}}. \bibinfo{pages}{9492--9503}.
\newblock


\end{thebibliography}

\appendix

\end{document}